\documentclass[11pt,letterpaper]{article}

\usepackage{times}
\usepackage{epsfig}
\usepackage{graphicx}
\usepackage{amsmath}
\usepackage{amssymb}
\usepackage{algorithmic}
\usepackage{algorithm}
\newcommand{\z}{z}

\begin{document}

\title{What is the Best Feature Learning
Procedure in Hierarchical Recognition Architectures?}

\author{Kevin Jarrett\\
\and
Koray Kvukcuoglu\\
\and
Karol Gregor\\
\and
Yann LeCun\\
Courant Institute of Mathematical Sciences, NYU\\
715 Broadway New York, NY 10003 \\
}

\date{}

\maketitle

\begin{abstract}

\noindent \textbf{NOTE: This paper was written in November 2011 and never published. It is posted on arXiv.org in its original form in April 2016.}\\
Many recent object recognition systems have proposed using a two phase
training procedure to learn sparse convolutional feature hierarchies: unsupervised
pre-training followed by supervised fine-tuning. Recent results suggest that these methods 
provide little improvement over purely supervised systems when the 
appropriate nonlinearities are included. This paper presents an empirical 
exploration of the space of learning procedures for sparse convolutional networks to assess
which method produces the best performance. 
In our study, we introduce an augmentation of the Predictive Sparse Decomposition 
method that includes a discriminative term (DPSD). We also introduce a new single phase supervised learning procedure that places an L1 penalty on the output state of each layer of the network. This forces the 
network to produce sparse codes without the expensive pre-training phase. Using DPSD with a new, complex predictor that incorporates lateral inhibition, combined with multi-scale feature pooling, and supervised refinement, the system achieves a 70.6\% recognition rate
on Caltech-101. With the addition of convolutional training, a 77\% recognition was obtained on the CIfAR-10 dataset. The surprising result is that networks using our purely supervised method with a simple encoder and multi-scale pooling produced competitive results, 
while being significantly simpler and faster to train.  This suggests that even task-oriented pre-training 
may not provide sufficient improvement to justify its use. 
We conclude with a set of experiments that show an important role of the local contrast
normalization in object recognition. We show that including the
nonlinearity preserves more information about the input in the output
feature maps leading to better discriminability among object
categories.

\end{abstract}

\section{Introduction}
Over the past few years there has been considerable interest in
learning sparse convolutional features for hierarchical recognition
architectures, such as feed-forward convolutional networks (or
ConvNets)~\cite{lecun-98, jarrett2009, koray-smooth, ranzato-nips-07,
  ranzato-cvpr10}, {\em deconvolutional} networks\cite{fergus-2010},
and various forms of Restricted Boltzmann Machines stacked to form
convolutional feature extractors~\cite{lee-nips-07,lee-icml-09}.  Most
of these authors advocate the use of a purely unsupervised phase to
train the features, optionally followed by a global supervised
refinement. However, a surprising set of results in~\cite{jarrett2009}
suggests that, on dataset with very few labelled samples such as
Caltech-101, learning the features (unsupervised or supervised) seems
to bring a relatively modest improvement over using completely random
features, provided that a suitable set of non-linearities is used at
each stage in the system (local mean removal and contrast
normalization). Other results suggest that when the number of labelled
samples grows, the unsupervised pre-training phase doesn't bring a
significant improvement over a purely supervised procedure.
Furthermore, while the learned filters at the first level seem fairly
generic and task independent (oriented Gabors), it is unclear whether
the mid-level features would not produce better results by being
pre-trained in a somewhat more task-specific manner.

Some recent works have advocated the use of sparse auto-encoders to
pre-train feature extractors in an unsupervised
manner~\cite{jarrett2009, koray-smooth}. In their Predictive Sparse
Decomposition method (PSD), and its convolutional version, a
non-linear feed-forward encoder function is trained to produce
approximations to the sparse codes that best reconstruct the
inputs under a sparsity penalty.

This paper startes with this basic approach and presents an empirical exploration of the space of 
learning procedures for sparse convolutional networks to assess which method produces the best performance. 
We study 1) a newly introduced discriminative version of PSD (DPSD), 2) a newly introduced single phase supervised
 training method, 3) convolutional versus patch based unsupervised learning, 4) the effect of using complex encoders
 that better predict the sparse code, 5) the use of a new multiresolution pooling method, 
and finally 6) contrast normalization's role in preserving information about about the input in the high level feature maps. 

Results are reported on Caltech-101 that are close to the best
methods that use a single feature type. Record results are reported on the CIfAR-10
dataset.

In order to study the effectiveness of task-oriented sparse coding methods,  we introduce 
an augmentation of the PSD  ~\cite{jarrett2009, koray-smooth} algorithm that includes a discriminative 
term in the objective function (DPSD). This allows us to directly evaluate the contribution of 
the discriminative term by comparing the performance of networks pretrained
with the respective algorithms. In order to study the effectiveness of pre-training altogether, we present 
a new purely supervised training method that places an L1 penalty on the 
output states of randomly initialized networks. This additional regularization drives 
the network towards producing sparse output maps and the resulting weight configuration 
resembles the structure that results from DPSD. We refer to this 
method as sparse-state supervised training. We further compare the traditional patch based
DPSD with a version of DPSD that is trained {\em convolutionally} over large image
regions~\cite{koray-smooth}, rather than on isolated patches. Training
convolutionally allows the system to take advantage of shifted
versions of all the filters to reconstruct images. This greatly
reduces redundancies and allows the resulting code to be much
sparser.  We also test two different predictors in order to determine how important accurate approximations to the optimal sparse codes are for performance. The simple encoder is the same used in ~\cite{jarrett2009}. 
We then introduce a new predictor architecture that incorporates a kind of lateral inhibition matrix following a linear filter bank and a shrinking function. This matrix is akin to the negative Gramm matrix of the filters 
used in the ISTA and FISTA algorithms for sparse coding [5]. This new encoder produces very accurate predictions [5], at the expense of additional computation. The encoder---pretrained or randomly initialized--- is followed by a rectification, a
local mean cancellation, a local contrast normalization, and a spatial
feature pooling. To augment the pooling stage, we introduce a new pooling scheme that
produces a multiresolution pyramid of features pooled over spatial neighborhoods of different sizes.
Lastly, we show that the effect of including the contrast normalization nonlinearity in an object recognition system is to preserve more information about the input image in the output feature
maps. This is counter-intuitive, as normalization could be thought of
as removing the total signal energy information.

\section{Learning Discriminative Dictionaries}

\noindent We consider the general model
\begin{equation}
\min_{\z} E(\z) \equiv H(\z) + G(\z)
\label{eq:main}
\end{equation}
\noindent where $H(.)$ is a smooth convex function that is
continuously differentiable and $G(.)$ is a continuous convex function
that is possibly nonsmooth. In recent years, there have been a number
of methods developed to solve this general model, such as the
Iterative Shrinking and Thresholding Algorithm
(ISTA)~\cite{daubechies}, or its accelerated extension
FISTA~\cite{beck-ista}. The $L_{1}$ regularization problem is a
popular special instance of this model where $H(\z) := ||x-D\z||^{2}$
and $G(\z) := ||\z||_{1}$.  In the case of classical sparse coding we
have the energy function:
\begin{equation}
  E(x,\z,D) = ||x- D\z||^{2}_{2} + \lambda_{1}||\z||_{1}
\label{eq:sp1}
\end{equation}
where $D\in \mathbb{R}^{m\times n}$ is the dictionary (generally
overcomplete), $x$ is input signal, $\z$ is the sparse code
representing $x$, and $\lambda_{1}$ is a hyperparameter controlling
the degree of sparsity. For a given input $x$, sparse coding finds the
optimal code $\z$ that minimizes $E(x,\z,D)$.  In classical sparse
coding, $D$ is hand picked for the task, e.g. (DCT, wavelet, or
Gabors)~\cite{Mallat:1993bs}, but the recent interest in sparse coding
revolves around the ability to learn an appropriate dictionary matrix
from data~\cite{Olshausen-Field}. Dictionary learning for image
feature extraction has recently led to state-of-the-art results in
image denoising~\cite{mairal-2008b} and image classification (see
~\cite{ylan-cvpr10, jarrett2009, mairal-cvpr-08, ranzato-nips-07} to
name a few).

Learning the dictionary allows us to adapt $D$ to the statistics of
the input signal, but it also allows us to enforce task-dependent
constraints~\cite{mairal2008sdl, mairal-cvpr-08, mairal-jm-2010,
  ylan-cvpr10}. If, for example, the objective is image classification
and labelled samples are available, then dictionaries can be learned
that are both reconstructive and discriminative. Basic reconstructive
sparse coding with a learned dictionary minimizes
\begin{equation}
  {\cal L}(D) = \sum_{i=1}^{m} \min_\z E(x^i,\z,D)
\label{eq:sp2}
\end{equation}
where $x^i, \; i=1\ldots m$ is a set of unlabelled training samples.
Considering the ability to enforce task dependence,
\cite{mairal2008sdl} proposed adding a discriminative term, ${\cal
  C}(y^{i},l(\z^{i},\theta))$, where $y^i$ is the label of sample
$x^i$, $l(.)$ is a (linear) discriminant function, ${\cal C}(.)$ is
some classification loss function, and $\z^i$ is the optimal sparse
code for $x^i$. Incorporating this term leads to the a redefined
energy function per sample $i$:
\begin{equation}
  E(x,y,\z,D,\theta) = {\cal C}(y,l(\z,\theta)) + ||x- D\z||^{2}_{2} + \lambda_{1}||\z||_{1}
\label{eq:sp4}
\end{equation}
The loss function for learning is
${\cal L}(D,\theta) = \sum_{i=1}^m \min_\z E(x^i,y^i,\z,D,\theta)$,
which is jointly minimized with respect to $D$ and $\theta$.

\section{Predictors}

In this section, we show how to use a trainable, non-linear,
feed-forward predictor to extend the use of discriminative
dictionaries to a larger class of recognition systems. In particular
we show how we can use this trainable module as a component in a
larger, globally-trained system with a flexible, multi-stage
architecture.

Following the PSD method~\cite{koray-psd-08, koray-smooth,
gregor-icml10} we use an encoder to approximate the optimal sparse
code, $\z^{*}$. This allows the computation of some $\hat{\z} \approx
\z^{*}$ with a rapid-feedforward pass through the encoder, instead of
using an iterative procedure to infer the optimal code. In
\cite{jarrett2009} the function used to approximate the $k$-th
component of the code was
\begin{eqnarray}
  F_{\tanh}(x)_k \equiv g_k \tanh \left( W_k . x + b_k\right)
\label{eq:tanh}
\end{eqnarray}
where $F_{\tanh}(x)_k$ is the $k$-th output, $g_k$ is a scalar gain,
$W_k$ is a the $k$-th dictionary element. The use of this function is
not particularly well justified, given that it makes it difficult to
produce outputs near zero. The only rationales seem to be that it is
similar to non-linear functions used in traditional neural nets, and
that it seems to yield easy convergence and reasonable recognition
performance.  

To address this, we also investigate the effect of using a richer encoder
based on the Iterative Shrinking and Thresholding Algorithm (ISTA) for sparse
coding ~\cite{daubechies,beck-ista} (see ~\cite{gregor-icml10} for details). The function
used to approximate the sparse code was

\begin{equation}
 F_{si}(x) \equiv sh\left(Wx - S . sh(W.x)\right)
\label{eq:si}
\end{equation}

\noindent The convolutional form is 
\begin{equation}
  F_{si}^p(x)  = sh\left( A_p - \sum_{q \ne p} S_{pq} . sh(A_q)\right)
\label{eq:si_conv}
\end{equation}
where $A_p = \sum_q W_{pq} * x_q$, $W_{pq}$ is the convolution kernel
connecting the $q$-th input feature map $x_q$ to the $p$-th output
feature map $F_{si}^p(x)$, $*$ is the convolution operator, and $S_{pq}$
is a (scalar) element of $S$.

Following~\cite{gregor-icml10}, instead of using the prescribed value
of $W$, $S$ and $b$, we will {\em train} these matrices and threshold
vectors so that the encoder best approximates the optimal sparse
code. However, to do so, it is preferable to make the shrinkage
function smooth and differentiable, so that gradients can be
back-propagated through it without trouble. Hence, we use the
soft-shrinkage function:
\begin{equation}
sh(x_{k}) = sgn(x_{k}) * (|x_{k}| + \frac{1}{\beta}\log(1 + e^{(\beta \times (b - |x_{k}|)))} - b ))
\end{equation} 
which is guaranteed to travel through the origin and is antisymmetric
\cite{koray-smooth}. The parameters $b$ and $\beta$ are learned from
data, and control the location of the smooth kink, and its kinkiness,
respectively.

\section{Training Methods}
\subsection{DPSD Optimization Procedure}

Unlike with the original PSD, the optimization procedure treats the
code prediction term separately. The optimal code is found by
minimizing $H(\z)+G(\z)$. Then, the encoder is adjusted to predict it
better. The linear discriminant function in the discriminative term of
equation~\ref{eq:fopt} is simply $l(\z,\theta) = u^{T}\z + r$, using
the notation $\theta = \{u,r\}$, where $u$ is a $c \times n$ matrix,
and $r$ a $c$-dimensional bias vector. The discriminative loss ${\cal
  C(.)}$ is the usual one for multinomial logistic regression: ${\cal
  C} = -l(z,\theta)_y + \log \sum_{y'=1}^c \exp(l(z,\theta)_{y'})$

The optimization occurs as block coordinate descent alternating
between the supervised sparse code update, and the update of
$D,W,b,\beta,\theta$. It should be noted that the columns of $D$ are
normalized to avoid trivial solutions. The pseudo-code for the
procedure is outlined in algorithm \ref{alg1}.
\begin{algorithm}[bt]                  
\caption{Task Oriented Predictors}         
\label{alg1}                             
\begin{algorithmic}                   
\STATE \textbf{input:} $(x^{i},y^{i})$ (sequence of signal pairs); $n$
(size dictionary); $\lambda_{i}$ (hyperparameters); niter (number
iterations); $F(.)$ (encoder type)
\STATE \textbf{output:} $D$, $W$, $\theta$
\STATE  \textbf{initialize:} $D$, $W$, $\theta$ to random values sampled from, 
 zero mean, unit stddev, Gaussian.
\FOR{$i = 0$ to niter}
\STATE \textbf{fix} $D = D^{i}$, $W = W^{i}$, $\theta = \theta^i$
\STATE \textbf{initialize} $\hat{\z}^{i} = F(W,x^{i})$ ~~ (initialize $\z$ with $\hat{\z}$) 
\STATE $\bullet$ Sparse Coding: solve
\STATE ~~~~~~~~~~~~$\z^{i*} = \arg \min_{\z} H(\z)+G(\z)$
\STATE $\bullet$ perform one step of stochastic gradient to minimize $||x^i - D\z^{i*}||^{2}_{2}$ with respect to $D$, and normalize.
\STATE $\bullet$ perform one step of stochastic gradient to minimize ${\cal C}(y^i,l(\z^{i*},\theta))$ with respect to $\theta$.
\STATE $\bullet$ perform one step of stochastic gradient to minimize $||\z^{i*} - F(x^i,W,b,\beta) ||^2_{2}$ with respect to $W,b,\beta$.
\ENDFOR
\end{algorithmic}
\end{algorithm}

The ISTA algorithm can easily be modified to account for the
discriminative term. Recall that ISTA (FISTA) solves the general model
given in equation \ref{eq:main}, where $H(\z)$ is smooth and convex and
$G(\z)$ is convex but possibly nonsmooth.  In this case we set:
\begin{equation}
  H(\z):={\cal C}(y,l(\z,\theta)) + \lambda_{1}||x - D\z||^{2}_{2} 
\label{eq:fopt}
\end{equation}
which is the sum of two convex functions and therefore convex
itself. We then have $G(\z):=||\z||_{1}$ as before which satisfies the
general models constraints. We can now use the FISTA algorithm to
infer the optimal sparse, discriminative code for a fixed input pair
$(x^i,y^i)$, dictionary $D$, and supervised parameter set
$\theta_{s}$.

This modification to $H(\z)$ forces the algorithm to discover a sparse
solution, $\z^{*}$ that strikes a balance between minimizing the
reconstruction error and maximizing class separability. Once the
supervised sparse code $\z^{*}$ for sample $i$ converges, its value is
fixed and a stochastic gradient step over $D$, $W$, $\beta$, $b$, and
$\theta$ is done. The process is iterated until the convergence of
$W$.

\subsection{Sparse-State Supervised Training}

\begin{figure}[bt]
  \centering 

   \includegraphics[width=1\linewidth]{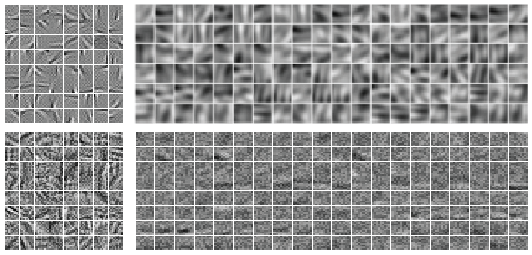}
  \caption{Top Left: first stage filters trained of a $F_{si}$ encoder
    trained with discriminative predictive sparse decomposition, using
    FISTA. Top Right: second stage filters trained the same way. Bottom Left:
    first layer filters in a randomly initialized network trained supervised with
    with an L1 penalty on the output states. Bottom Right: second stage filters
    trained the same way. Although the supervised filters are noisier, it is clear
    they learn structure similar to the unsupervised filters. }
\label{cltech:filt}
\vspace{-0.5cm}
\end{figure}

This section introduces a new purely supervised training
procedure that places an L1 penalty on the output states. This additional 
regularization drives the network towards weight configurations that produce sparse 
output maps. The method is somewhat similar to DPSD as the training involves 
the integration of signals from the sparsity penalty and from the discriminative loss function, however, 
the training is performed in a single supervised phase eliminating the costly unsupervised phase. 
The loss function is now ${\cal C} = -l(z,\theta)_y + \log \sum_{y'=1}^c \exp(l(z,\theta)_{y'}) + \lambda_{L1}||z||_{1}$. 
This additional regularization trains the network to produce sparse codes without requiring an expensive iterative routine 
for inference. This \textit{dramatically} reduces the overall training time while producing codes similar to those produced 
using DPSD. The L1 penalty can be assessed at any stage of the network, such as just after the encoder or after the pooling. We found 
that assessing the penalty after the pooling produced the best performance.  This also introduces an additional 
parameter: the L1 penalty weight, $\lambda_{L1}$. This is determined by cross-validation and we found $\lambda_{L1}=0.4$ 
produced the best results. This training procedure produces noisy edge detectors at the first layer and filters similar 
to those produced by DPSD for the second layer. The filters are shown in figure \ref{cltech:filt}

\section{Constructing Multi-Stage Systems}

The encoders trained with this approach can be used in multi-stage
object recognition architectures, similar to the ConvNets as proposed
in ~\cite{jarrett2009,koray-smooth}.

Each stage is composed of an encoder, a rectifier, a local contrast
normalization, and a pooling operation. There are two types of
encoders, one denoted by $F_{si}$ and described by equation
\ref{eq:si_conv}, and one denoted by $F_{tanh}$ and described by
equation \ref{eq:tanh}. The encoders can be trained with or without
the discriminative term.  The filters \textit{of the first stage}
encoder may be pre-trained at the patch level, or {\em
  convolutionally} over larger image windows. However, \textit{none of
  the results reported here use convolutional pre-training for the
  second stage}. The encoder module is followed by a pointwise
absolute value rectification module denoted by $R_{abs}$, a local mean
removal and contrast normalization module denoted by $N$. This module
is identical to that of~\cite{jarrett2009, koray-smooth}: each value
is replaced by itself minus a Gaussian-weighted average of its
neighbors over a $9\times 9$ spatial neighborhood, standard deviation
$1.6$, and over all feature maps. Then, each value is divided by the
standard deviation of its neighbors over the same neighborhood, or by
a constant, whichever is largest. The final step of a stage is a
pooling module, that is either a max pooling, denoted by $P_{M}$, or
an average pooling denoted by $P_{A}$. The pooling module computes the
max or average of the values in a each feature map over a pooling
window. The window is then stepped by a given stride over the feature
map (horizontally and vertically) to produce the neighboring
values. The stride determines a downsampling ratio for the pooling
layer.

Multiple stages can be stacked on top of each other, each stage taking
as its input the output of the previous stage. For the experiments, we
used single-stage and dual-stage systems. In this work, we also
experimented with a new pooling scheme for the second stage, called
pyramid average pooling and denoted by $P_{A}^{pyr}$. The idea is to
use multiple pooling window sizes and downsampling ratios so as to
produce multiple version of the top feature map at several
resolutions. The rationale is to produce a spectrum of representations
between spatially organized features and global features (akin to bags
of features).  This is reminiscent of the spatial pyramid pooling of
Lazebnik et al.~\cite{lazebnik-cvpr-06}.

Finally, the output of the last stage is fed to a multinomial logistic
regression classifier trained in supervised mode with the
cross-entropy loss. When global supervised refinement is performed,
the gradient of the loss is back-propagated down the entire
architecture. All the parameters (all filters, $S$ matrices, biases,
etc) can be updated with one step of stochastic gradient descent.

\section{Experiments with Caltech-101}

For preprocessing, the images were converted to grayscale and
downsampled to $151\times151$, then contrast-normalized with a local
$9\times9$ Gaussian-weighted neighborhood to remove the local mean and
scale the neighborhood to unit std producing
$143\times143$ input images.

\subsection{Architecture} 

\noindent \textbf{Stage One:} The first stage consist of an
$F_{\tanh}$ or $F_{si}$ encoder module with 64 filters of size
$9\times9$ followed by a rectification module ($R_{abs}$, local mean
removal and contrast normalization module ($N$), and average pooling
module $P_{A}$ with $10\times10$ pooling window and $5\times5$
downsampling. The output of this stage consists of $64$ feature maps
of size $26\times26$. This output is used as the input to the second
stage.
\begin{table}[bt]
\label{cltech}
\begin{centering}
  \begin{tabular}{|l|l|c|}
\hline
\multicolumn{2}{|c|}{\bf SIFT-SparseCoding-Pooling-SVM}     & \textbf{$\%$}  \tabularnewline
\hline
\multicolumn{2}{|c|}{\bf Lazebnik etal ~\cite{lazebnik-cvpr-06}}    & $64.6\pm.08$   \tabularnewline
\multicolumn{2}{|c|}{\bf Yang etal ~\cite{yang-cvpr-09}}       & $73.2\pm0.5$   \tabularnewline  
\multicolumn{2}{|c|}{\bf Boureau etal ~\cite{ylan-cvpr10}}       & $75.7 \pm 1.1$   \tabularnewline  
\hline  
\hline  
\multicolumn{2}{|c|}{\bf ConvNets variants (learned features)}     & \textbf{$\%$}  \tabularnewline
\hline  
\multicolumn{2}{|c|}{\bf Jarrett etal ~\cite{jarrett2009}}    & $65.6\pm1.0$   \tabularnewline
\multicolumn{2}{|c|}{\bf Kavukcuoglu etal ~\cite{koray-smooth}}  & $66.3\pm1.5$   \tabularnewline
\multicolumn{2}{|c|}{\bf Lee etal~\cite{lee-icml-09}}          & $65.4\pm0.5$   \tabularnewline
\multicolumn{2}{|c|}{\bf Zeiler etal ~\cite{fergus-2010}}       & $66.9\pm1.1$   \tabularnewline
\multicolumn{2}{|c|}{\bf Ahmed etal ~\cite{ahmed-nec}}       & $67.2\pm1.0$   \tabularnewline   
\hline
\hline
\textbf{Architecture} & \textbf{Protocol} & \textbf{$\% $}\tabularnewline
\hline
\textbf{(1) $F_{\tanh}-R_{abs}-N-P_{A}^{pyr}$} & $\mathbf{R^{+}R^{+}}$  & $65.4 \pm 1.0$   \tabularnewline
\textbf{(2) $F_{\tanh}-R_{abs}-N-P_{A}^{pyr}$} & $\mathbf{R^{+}_{L1}R^{+}_{L1}}$  & $\textbf{69.0} \pm \textbf{0.2}$   \tabularnewline
\textbf{(3) $F_{\tanh}-R_{abs}-N-P_{A}^{pyr}$} & $\mathbf{U^{+}U^{+}}$  & $66.2 \pm 1.0$   \tabularnewline
\hline
\textbf{(4) $F_{si}-R_{abs}-N-P_{A}$} & $\mathbf{R^{+}R^{+}}$  & $63.3 \pm 1.0$   \tabularnewline  
\textbf{(5) $F_{si}-R_{abs}-N-P_{A}$} & $\mathbf{R^{+}_{L1}R^{+}_{L1}}$  & $64.3 \pm 1.0$   \tabularnewline  
\textbf{(6) $F_{si}-R_{abs}-N-P_{A}$} & $\mathbf{UU}$  & $60.4 \pm 0.6$   \tabularnewline  
\textbf{(7) $F_{si}-R_{abs}-N-P_{A}$} & $\mathbf{U^{+}U^{+}}$  & $66.4 \pm 0.5$   \tabularnewline  
\textbf{(8) $F_{si}-R_{abs}-N-P_{A}^{pyr}$} & $\mathbf{U^{+}U^{+}}$  & $67.8 \pm 0.4$   \tabularnewline  
\textbf{(9) $F_{si}-R_{abs}-N-P_{A}$} & $\mathbf{DD}$  & $66.0 \pm 0.3$   \tabularnewline  
\textbf{(10) $F_{si}-R_{abs}-N-P_{A}$} & $\mathbf{D^{+}D^{+}}$  & $68.7\pm 0.2$   \tabularnewline
\textbf{(11) $F_{si}-R_{abs}-N-P_{A}^{pyr}$} & $\mathbf{D^{+}D^{+}}$  & $\textbf{70.6} \pm \textbf{0.3}$   \tabularnewline    
\hline
\end{tabular}
\vspace{0.0in}
\caption{Recognition accuracy on the Caltech-101 dataset using various
  architectures and training protocols. $F_{tanh}$ indicates an
  encoder with a $\tanh$ non-linearity followed by a gain matrix,
  while $F_{si}$ indicates a soft-shrink non-linearity followed by a
  trained lateral inhibition matrix $S$. Pyramid average pooling at
  the penultimate layer is denoted by $P_A^{pyr}$.  $R$ indicates
  randomly initialized filters, $R_{L1}$ is random initialization with the 
  additional L1 penalty, $U$ unsupervised pre-training, and $D$
  pre-training with a discriminative term. Superscript ``+'' indicates
  global supervised fine-tuning.}
\end{centering}
\vspace{-0.23in}
\end{table}
\noindent \textbf{Stage Two:} The second stage is composed of a
$F_{\tanh}$ or $F_{si}$ encoder module with 256 output feature maps,
each of which combines a randomly-picked subset of 16 feature maps
from the first stage using $9\times9$ kernels. The total number of
convolution kernels is $256\times 16 = 4096$. The size of the feature
maps (before pooling) is $18\times 18$. Two types of pooling modules
were used: average pooling, and pyramid average pooling. The average
pooling used $6\times6$ with $4\times4$ downsampling, producing 256
feature maps of size $4\times 4$.

The pyramid average pooling uses multiple pooling window sizes and
downsampling ratios. In addition to the 256x4x4 feature maps produced
by 6x6 and 4x4 downsampling mentioned above, we added 256x3x3 feature
maps produced by 8x8 pooling with 5x5 downsampling, 256x2x2 feature
maps produced by 10x10 pooling with 8x8 downsampling, and 256x1x1
feature maps produced by 18x18 pooling. The results of the various
pooling sizes are concatenated and fed to the classifier at the next
layer.

\noindent \textbf{Classifier Stage:} The Stage 2 output (a single set
of 256 feature maps, or multiple sets in the case of pyramid pooling)
is fed to a classifier stage, which consists in an L1-L2-regularized
multinomial logistic regression classifier trained with the
cross-entropy loss using stochastic gradient descent. The
hyperparameters for the unsupervised and supervised phases of 
training were selected by cross validation.

\subsection{Results} 
Table 1 summarizes the results for various architectures and training
protocols. We also include other published results for comparison. The
methods listed use a framework similar to the one presented here.  All
results are reported as the average accuracy over 5 random splits of
the dataset, each with 30 training samples per category.  The training
protocols are denoted by one letter for single-stage systems and two
letters for two-stage systems. The letter $R$ denotes random filters,
$U$ denotes unsupervised pre-training, and $D$ pre-training with a
discriminative term. Superscript $+$ indicates global training using
supervised gradient-descent. Comparisons between the various
architectures and training protocols are discussed below.  The use of
max pooling instead of average pooling was attempted, but resulted in
a slight decrease in accuracy of $(\approx 1\%)$.

\noindent \textbf{1.} Experiments (8) and (11) show that including the
discriminative term in the unsupervised pre-training increases the
performance by $\approx 3\%$. Our best method reaches $70.6\%$, which
compares favorably with other methods in which the entire feature
hierarchy is trained. Methods with better accuracy~\cite{yang-cvpr-09,
ylan-cvpr10} use SIFT at the first stage and use sparse coding with a
large code size at the second stage.

\noindent \textbf{2.} Surprisingly, sparse-state supervised training
with random initialization achieves $69\%$ with the simpler, $F_{\tanh}$
encoder. This is $1.7\%$ off of our best method with \textit{considerably} less
computation. This method is \textit{significantly} simpler than every method that outperforms it
and only requires a single stage of training. The poor performance, $64.3\%$, while using the 
$F_{si}$ encoder is likely due to convergence issues with the random initialization of this encoder.

\noindent \textbf{3.} Comparing network $(9,6)$ shows that in the
absence of supervised fine-tuning, including the discriminative term increases
the performance of the network by $6\%$.

\noindent \textbf{4.} It is clear from comparing (9,7) that including
a discriminative term, either during the unsupervised phase or after
(as fine-tuning), results in the same performance. The advantage of
including the term in the unsupervised phase is that we can fold all
the learning into a single phase, as opposed to two separate phases.

\noindent \textbf{5.} Comparing (9,10) shows that even though we have
the discriminative term in the unsupervised phase, adding fine-tuning
still increases the performance. This is because we have global
training with a hierarchical system so that the second layer is able
to effect the output feature maps of the first layer. This creates
room for improvement which here is $2.7\%$.

\noindent \textbf{6.} A comparison of networks $(7,8)$ and $(10,11)$,
shows that the pyramid method does increase the performance. For the
full network with the discriminative term the increase is nearly
$2\%$.

\section{Experiments with CIfAR-10}

The CIfAR-10 dataset ~\cite{cifar} is a hand-labeled subset of a
larger dataset of 80 million $32\times32$ color images from the
web. There are $50,000$ images, $5,000$ per class,
with $1,000$ images per class set aside for testing. The low
resolution and high level of variability make this a challenging
dataset for most recognition systems. 

We use the following protocol: The data is composed of $32\times32$
RGB color images. For preprocessing, the images were converted to YUV
(luminance-chrominance). The Y (luminance) channel was locally
normalized (with an $N$ module), and the U and V chrominance channels
had their global mean subtracted off, and were divided by their global
standard deviations.

\begin{figure}
\centering
  \includegraphics[width=0.8\linewidth]{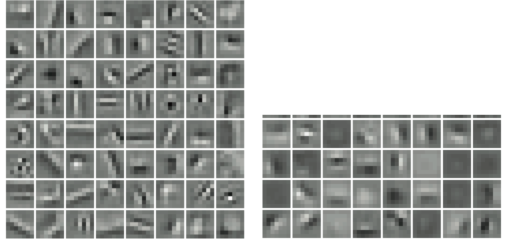}
\caption{First stage $7\times 7$ filters learned convolutionally on
  $16\times16$ color image patches (YUV encoded). The left side shows
  the 64 filters for the Y luminance channel. The right side shows the
  32 filters for the U and V chroma channels. These chroma filters are
  lower frequency than the Y filters.}
\label{fig:cifar1}
\vspace{-0.25in}
\end{figure}
\subsection{Architecture}

\noindent \textbf{Stage One:} Results are reported for two separate
pipelines. The only differences between the two is the pooling
operator (average or max) and whether the normalization layer comes
before or after the pooling layer. The best performing network uses
max-pooling with an encoder $F_{si}$ trained convolutionally with
$7\times7$ kernels on $16\times16$ image patches. There are 64 output
feature maps and 96 convolution kernels. Y channel (luminance) is
connected to all 64 output feature maps and each one of U and V
channels is connected to randomly chosen 16 output feature maps. This
is followed by an absolute value rectification, a max-pooling over
$4\times4$ windows with with a $2\times2$ downsampling, and a local
mean removal and contrast normalization with a $9\times9$ window.  The
average pooling network is identical, except that order of the
normalization and pooling operations is reversed. The output of this
layer is $64\times12\times12$.

It was found empirically that average pooling produces better results
when {\em preceded} by the normalization module, while max-pooling is
better when {\em followed} by the normalization module.  One possible
explanation is that gradient-based supervised learning converges
faster and to a better solution when the states of the layers have
approximately zero mean and uniform variances. The contrast
normalization produces zero-mean and unit variance states. The average
pooling modules does not affect the mean, and scales all the variances
by the pooling window size. By contrast, the max pooling shifts the
means to positive values, and reduces the variances unequally. Placing
the normalization module after the max pooling will restore the state
to zero mean and uniform variance.

The first stage filters are shown in figure~\ref{fig:cifar1}. The
filters on the Y channel are somewhat more varied than the Caltech-101
filters, because they have been trained convolutionally. Convolutional
training avoids the appearance of shifted versions of each filters,
since shifted versions are already present due to the
convolutions. This causes filters to be considerably more diverse, as
shown by~\cite{koray-smooth}. The filters to the right are from the U
and V channels. The filters exhibit a clear structure, though they are
considerably low frequency than the Y filters.

\noindent \textbf{Stage Two:} The output of the 1st stage is the input
to the 2nd. The encoder uses $7\times7$ kernels on $7\times7$ image
patches (patch-based, non-convolutional training). The second stage
has 256 output feature maps, each of which combines a random subset of
32 feature maps from the previous stage for a total of 8192
convolution kernels. This is followed by an absolute value
rectification, a max pooling with $3\times3$ window without
downsampling, and a local contrast normalization with a $3\times3$
windows.  The average pooling network is identical, but with the order
of the normalization and pooling operations reversed. The output of
this stage is $256\times4\times4$. The hyperparameters for the
supervised and unsupervised training were selected by cross
validation. 

\subsection{Results}

\begin{table}[bt]
\begin{centering}
  \begin{tabular}{|l|c|c|}
\hline
\multicolumn{2}{|c|}{\bf Method}     & $\%$  \tabularnewline
\hline
\hline  
\multicolumn{2}{|l|}{\bf $10000$ linear combinations \cite{ranzato-cvpr10}} & $36.0\%$   \tabularnewline 
\multicolumn{2}{|l|}{\bf $10$k GRBM, 1 layer with fine-tuning \cite{cifar}} & $64.8\%$   \tabularnewline 
\hline
\multicolumn{2}{|l|}{\bf mcRBM-DBN(11025-8192-8192) \cite{ranzato-cvpr10}} & $71.0\%$   \tabularnewline  
\multicolumn{2}{|l|}{\bf PCA(512)-iLCC(4096)-SVM \cite{yu-icml10}} & $\textbf{74.5}\%$  \tabularnewline
\hline
\textbf{Architecture} & \textbf{Protocol} & \textbf{$\% $}\tabularnewline
\hline
\textbf{(1) $F_{si}-R_{abs}-P_{M}-N$} & $\mathbf{RR}$  & $47.5\%$   \tabularnewline
\textbf{(2) $F_{tanh}-R_{abs}-P_{M}-N$} & $\mathbf{R^{+}R^{+}}$  & $70.0\%$   \tabularnewline
\textbf{(3) $F_{si}-R_{abs}-P_{M}-N$} & $\mathbf{R^{+}R^{+}}$  & $70.5\%$   \tabularnewline  
\textbf{(4) $F_{tanh}-R_{abs}-P_{M}-N$} & $\mathbf{R^{+}_{L1}R^{+}_{L1}}$  & $\textbf{73.1\%}$   \tabularnewline

\hline
\textbf{(5) $F_{si}-R_{abs}-P_{M}-N$} & $\mathbf{D_{c}^{+}}$  & $59.6\%$   \tabularnewline  
\textbf{(6) $F_{si}-R_{abs}-N-P_{A}$} & $\mathbf{D_{c}^{+}}$  & $60.0\%$   \tabularnewline  
\hline
\textbf{(7) $F_{tanh}-R_{abs}-P_{M}-N$} & $\mathbf{U_c^{+}U^{+}}$  & $74.7\%$   \tabularnewline
\textbf{(8) $F_{si}-R_{abs}-P_{M}-N$} & $\mathbf{U_c^{+}U^{+}}$  & $74.8\%$   \tabularnewline  
\textbf{(9) $F_{si}-R_{abs}-P_{M}-N$} & $\mathbf{D^{+}D^{+}}$  & $74.4\%$   \tabularnewline  
\textbf{(10) $F_{si}-R_{abs}-N-P_{A}$} & $\mathbf{D_{c}^{+}D^{+}}$  & $75.0\%$   \tabularnewline  
\textbf{(11) $F_{si}-R_{abs}-P_{M}-N$} & $\mathbf{D_{c}^{+}D^{+}}$  & $\textbf{77.6\%}$   \tabularnewline    
\hline
\end{tabular}
\vspace{.03in} 
\caption{Recognition accuracy on the CIfAR-10 dataset for different
  architectures, and training procedures. Notations are as in Table
  1. Subscript $c$ indicates convolutional pre-training (as opposed to
  patch-based).}
\end{centering}
\vspace{-0.21in}
\end{table}

Table 2 shows our method with the current state of the art in the
published literature. The effects of various components is broken down
in the points below. First of all, our best method yields $77.6\%$
correct recognition, which is the best ever reported in the
peer-reviewed literature\footnote{a recent unpublished report
  from U. of Toronto also reports 77.6\% using a convolutional Deep
  Belief Net, (A. Krizhevsky, October 2010).}

\noindent \textbf{1.} Lines (8) and (11) show that including the
discriminative term increases the performance by about $3\%$. This is
similar to the increase seen with Caltech-101. Unsupervised
pre-training makes a 4\% difference over random initialization and
purely supervised training, with convolutional, or non-convolutional
discriminative. But the jump is 7\% when the pre-training is both
convolutional and discriminative.

\noindent \textbf{2.} Comparing (4,9) we again see that sparse state
supervised training is competitive with patch based discriminative pre-training.
The difference grows when convolutional training is used at the first stage (11).
 We note that this increase in performance from patch to convolutional training was not seen with 
 training Caltech 101. 
 
\noindent \textbf{3.}  There is a considerable jump of nearly $3\%$
(lines (9) and (11)) obtained from convolutional training. It seems
that because the internal feature representation in this CIfAR network
has lower dimension (lower spatial resolution), the redundancy
reduction afforded by the convolutional training makes a significant
difference.

\noindent \textbf{4.} Consistent with the findings of
\cite{ylan-cvpr10}, comparing (10) and (11) shows that max pooling
outperforms average pooling by $2.6\%$. It is surprising that we have
not seen this effect with Caltech. Perhaps the high level of variation
in Cifar is the reason that max pooling outperforms average on this
dataset. It may also be that blurring the feature maps when they are
already at such a low resolution removes information necessary to
learn good features.

\vspace{-0.11in}
\section{Effects of Contrast Normalization}
\vspace{-0.05in}
\begin{figure}
\centering
  \includegraphics[width=0.9\linewidth]{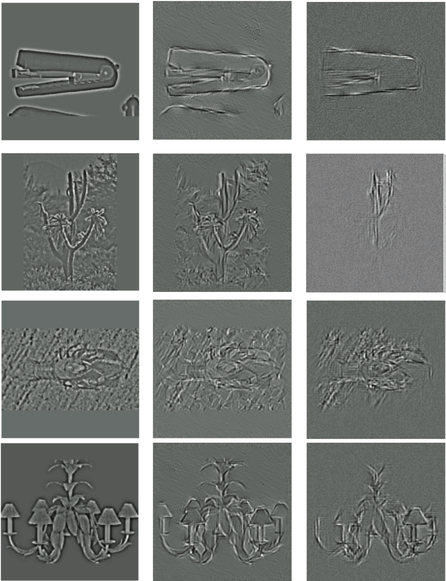}
  \caption{Hallucinated images that produce a given feature
    representation.  Images from the Caltech-101 dataset (shown on the
    top row) were ran through a two-stage trained ConvNet. The output
    feature maps were recorded. Hallucinated inputs (middle and bottom
    rows) were produced by performing steepest descent in input space
    from a random initial condition so as to produce the same feature
    maps as the original images. The middle row shows reconstructions
    for a network with contrast normalization modules ($N$) at both
    stages, and the bottom row for a network without contrast
    normalization.}
\label{fig:hallu}
\vspace{-0.21in}
\end{figure}

This section looks at the impact of including the local contrast
normalization nonlinearity (CN) in the feature extraction process. CN
has been applied to areas as diverse as sensory
processing~\cite{heeger-92, simoncelli-96},
image denoising~\cite{portilla}, and redundancy
reduction~\cite{div-norm} to name a few. It is also
known to improve performance in higher level tasks such as object
recognition \cite{jarrett2009}, though the reasons for this are poorly
understood.

The networks described in this section are the same as those used in
the Caltech 101 experiments with 3 important differences: 1) the
second layer is 128 instead of 256; 2) the number of connections
between the two layers is 32 instead of 16; 3) the pooling and
downsampling ratios are decreased.

The first layer has 64 $9\times9$ convolution kernels followed by an
absolute value rectification, local CN (or not), and $5\times5$
pooling window with a $2\times2$ downsampling step size. The results
in feature maps that are $64\times66\times66$. These feature maps are
the input to the second layer.  The second layer has 128 output
feature maps that combine a random subset of 32 feature maps from the
first layer using $9\times9$ kernels. This is followed by an absolute
value rectification, local CN (or not), and a $4\times4$ pooling
window with a $2\times2$ downsampling step size. This results in
output feature maps that are $128\times28\times28$.

The experiments are done on entire images from Caltech 101. First an
image is projected through the 2 layer network to produce a set of
feature maps that represent the input in feature space. These feature
maps are fixed and the input image is randomized. Steepest descent
optimization is then used to find the optimal input for that
particular set of feature maps. This process is carried out with and
without CN.

The results for four different images randomly drawn from Caltech 101
are shown in figure~\ref{fig:hallu}. The top row is the original image
which has been preprocessed. The center row is the optimal input found
when CN is included. The bottom row is the optimal input found without
CN. It is clear that the images synthesized using CN are are more
similar to the original input images.
\begin{table}[bt]
\centering
\begin{tabular}{|l|c|c|}
\hline
Network & Protocol & Accuracy $\%$ \tabularnewline
\hline
\bf{$F_{si}-R_{abs}-N-P_{A}$} & $\mathbf{DD}$ & $64.1\%$\tabularnewline
\bf{$F_{si}-R_{abs}-P_{A}$} & $\mathbf{DD}$ & $59\%$\tabularnewline
\hline
\end{tabular}
\vspace{.03in}
\caption{The recognition accuracy for the networks used to synthesize
  the hallucinated images in figures~\ref{fig:hallu}.}
\vspace{-0.21in}
\end{table}

This suggests that one way CN improves recognition performance is by
preserving more information about the input in the feature maps. These
richer feature maps provide the classifier with important
characteristics of the object class that help to increase
discriminability. Table 3 shows the performance of the networks on
Caltech 101 $(UU)$. The increase in the networks ability to accurately
synthesize the input image is correlated with an increase in its
performance on the recognition task.

\vspace{-0.09in}
\section{Discussion}
\vspace{-0.05in}

This paper presents a thorough exploration of various training 
procedures for sparse, convolutional feature hierarchies. In our analysis,
we introduce 1) a new unsupervised learning algorithm (DPSD) that includes a 
discriminative term in the sparse coding objective function, 2) a new single phase
supervised training procedure that produces results similar to two-phase training
methods, 3) a new multiresolution pooling mechanism that reliably increases 
recognition performance over traditional pooling, and 4) a new nonlinear feedforward encoder that can be trained to approximately predict
the optimal sparse codes by using a smooth shrinkage nonlinearity and a
cross-inhibition matrix. Additionally, we have shown that
convolutional training of this architecture also improves
performance. We tested this system on Caltech-101 and CIfAR-10, producing
state of the art or comparable results on both datasets. The surprising result 
is that discriminative unsupervised pre-training with a complex encoder works well, but pure supervised with sparsity is almost as good and much simpler. We feel that these results make the use 
of complex multi-stage pre-training procedures unnecessary unless convolutional learning
proves helpful. Finally we provided evidence indicating that contrast normalization improves
recognition performance by preserving more information about the input
necessary for discrimination in the feature maps.

 \vspace{-0.1in}
\renewcommand{\baselinestretch}{0.9}
{\small
\bibliographystyle{ieee}
\bibliography{total}
}

\end{document}